\documentclass[letterpaper, 10 pt, conference]{ieeeconf}

\IEEEoverridecommandlockouts

\usepackage{cite}
\usepackage{newtxtext}
\usepackage{mathptmx}
\usepackage[T1]{fontenc}
\usepackage{amsmath,amssymb,amsfonts}
\usepackage{graphicx}
\usepackage{textcomp}
\usepackage{amssymb}
\usepackage{multirow}
\usepackage{multicol}
\usepackage{float}
\usepackage{url}
\usepackage{booktabs}
\usepackage{adjustbox}
\usepackage{algorithm}
\usepackage{algpseudocode}
\usepackage{threeparttable}
\usepackage{pdflscape}
\usepackage{graphicx} 
\usepackage{array} 
\usepackage{xcolor}
\definecolor{lightblue}{rgb}{0.68, 0.85, 0.9} 
\def\BibTeX{{\rm B\kern-.05em{\sc i\kern-.025em b}\kern-.08em
    T\kern-.1667em\lower.7ex\hbox{E}\kern-.125emX}}

\title{\LARGE \bf
Attention Mamba: Time Series Modeling with Adaptive Pooling Acceleration and Receptive Field Enhancements}

\author{
Sijie Xiong$^{1}$, Shuqing Liu$^{1}$, Cheng Tang$^{1}$, Fumiya Okubo$^{1}$, Haoling Xiong$^{2}$, and Atsushi Shimada$^{1}$
\thanks{$^{1}$Graduate School of Information Science and Electrical Engineering, Kyushu University, W2-828, 744, Motooka, Nishi-ku, Fukuoka, 819-0395 Japan.
{\tt\small xiong.sijie.630@s.kyushu-u.ac.jp}}
\thanks{$^{2}$Graduate School of Business, ECUST, East China University of Science and Technology, 130 Meilong Road, Shanghai, 200237, China.}
}

\begin{document}

\maketitle
\thispagestyle{empty}
\pagestyle{empty}
\begin{abstract}
"This work has been submitted to the lEEE for possible publication. Copyright may be transferred without noticeafter which this version may no longer be accessible." 

Time series modeling serves as the cornerstone of real-world applications, such as weather forecasting and transportation management. Recently, Mamba has become a promising model that combines near-linear computational complexity with high prediction accuracy in time series modeling, while facing challenges such as insufficient modeling of nonlinear dependencies in attention and restricted receptive fields caused by convolutions. To overcome these limitations, this paper introduces an innovative framework, Attention Mamba, featuring a novel Adaptive Pooling block that accelerates attention computation and incorporates global information, effectively overcoming the constraints of limited receptive fields. Furthermore, Attention Mamba integrates a bidirectional Mamba block, efficiently capturing long-short features and transforming inputs into the Value representations for attention mechanisms. Extensive experiments conducted on diverse datasets underscore the effectiveness of Attention Mamba in extracting nonlinear dependencies and enhancing receptive fields, establishing superior performance among leading counterparts. Our codes will be available on GitHub.

\end{abstract}

\section{INTRODUCTION}

Time series data, a collection of observations gathered at consecutive time intervals, construct the fundamentals of various real-world applications including traffic flow management, financial predictions, electricity consumption forecasts, healthcare supervisions, and more \cite{B01}. Time series modeling, hence, has attracted considerable attention from academia and industries, including long-term forecasting, short-term forecasting, and contrast forecasting \cite{B8, B18, B20}. Therefore, the development of robust models for time series forecasting (TSF) to uncover hidden patterns and predict the future states remains a critical area of research \cite{B1}. The core of such research mainly focuses on capturing temporal dependencies (TD) and inter-variate correlations (VC) \cite{B2}. 

During the early stages, Transformer-based models demonstrate exceptional capabilities in capturing both TD and VC, leading to remarkable effectiveness in TSF \cite{B8}. The encoder-decoder structure inherently offers Transformer-based models considerable advantages in parsing global and local patterns \cite{B15, B19, B20}. However, as TSF becomes increasingly complex, the attention mechanisms employed in both encoders and decoders pose significant challenges for Transformer-based models. These challenges are reflected in the exponentially increasing computational costs, driven by the explosion of data, parameters, and variates involved in complex scenarios \cite{B3}. To mitigate the quadratic growth in computational demands, lightweight Transformer-based models focus on processing a subset of sequences. While this approach enhances computational efficiency, it often comes at the expense of performance, as inadequate feature extraction leads to diminished accuracy \cite{ B04, B05}. On the other hand, Linear-based models prioritize fast attention mechanisms by utilizing standard linear layers to compute approximate attention efficiently \cite{B21, B23, B06}. However, relying exclusively on linear layers results in limited nonlinear dependency modeling and inadequate in-context feature extractions, rendering Linear-based models ineffective in handling complex scenarios \cite{B23}. As noted in \cite{B23}, Linear-based models can surpass Transformer-based models in simple scenarios. However, their performance declines significantly in complex situations or when input information is insufficient.

In recent studies, State Space Models (SSMs) have demonstrated performance comparable to Transformer-based models in terms of accuracy, while achieving a similar level of computational efficiency as Linear-based models \cite{B4}. By utilizing $V$ convolutional layers (Conv1D, as shown in Fig. \ref{figure1}(c)), SSMs effectively capture both linear and nonlinear dependencies, resulting in enhanced accuracy. Moreover, their computational efficiency is enhanced through the integration of parallel processing and state space representations \cite{B4, B5}. To further enhance training and inference speed, SSMs integrate hardware-aware selective mechanisms, giving rise to Mamba. Through hardware scanning, Mamba becomes input-dependent, dynamically adapting its state space representations. Meanwhile, the selective mechanism equips Mamba with both long-term and short-term memory capabilities \cite{B5}. Consequently, Mamba-based models have demonstrated exceptional effectiveness and impressive performance in TSF. For example, Simple-Mamba (S-Mamba) \cite{B3} utilizes a bidirectional Mamba block to enhance nonlinear dependency modeling, SiMBA \cite{B9} focuses on extracting features from frequency domains, and MixMamba \cite{B08} incorporates a mixer-of-experts mechanism to overcome knowledge bottlenecks. However, as suggested by the performance of Mamba-based models, which lags behind that of Transformer-based and Linear-based models, there is a pressing need for enhanced receptive fields. This improvement is crucial to better capture nonlinear dependencies and promote the overall performance of Mamba-based models \cite{B3, B5, B9, B08, B09, B010}.

To address the limited receptive fields, we introduce an innovative Mamba-based model, Attention Mamba (Fig. \ref{figure1}(a)). Initially, to empower Attention Mamba with global receptive fields and accelerate attention computations, we creatively design an attention mechanism, Adaptive Pooling (Fig. \ref{figure1}(b)), where \textbf{Query} and \textbf{Key} are down-scaled to one quarter by adaptive average and max pooling techniques. The reductions in the dimensions of \textbf{Query} and \textbf{Key} forms a faster computation while preserving global information. Besides, the adoption of adaptive pooling techniques can provide nonlinearity to fast attentions, mitigating the drawbacks of Linear-based models.  Moreover, to enhance the nonlinear long-short dependencies and provide \textbf{Value} to Attention Mamba, we introduce the bidirectional Mamba block (Fig. \ref{figure1}(a)), where the bidirectional mechanism can better refine critical features of original inputs. In summary, our designs enable Attention Mamba to widen the receptive fields and improve performance with enhanced nonlinear dependency extractions. Our contributions can be concluded as follows:
\begin{itemize}
    \item We creatively design an Adaptive Pooling block to enhance nonlinear dependency extractions in attentions and widen receptive fields by providing global features through adaptive pooling techniques, while also effectively alleviating the computational complexity of attention mechanisms.
    \item We propose a novel Mamba-based model, Attention Mamba, by integrating the previous Adaptive Pooling block with a bidirectional Mamba block. The two blocks boost refined features and achieve an equilibrium between computational complexity and accuracy, improving the comprehensive performance of Attention Mamba.
    \item Extensive experiments on benchmark models across diverse datasets are implemented, demonstrating the excellent performance of Attention Mamba among leading counterparts. We also discuss the advantages and drawbacks of Attention Mamba in the end.
\end{itemize}

\section{METHODOLOGY}
\begin{figure*}[!htbp]
\centering
\includegraphics[width=0.7\textwidth]{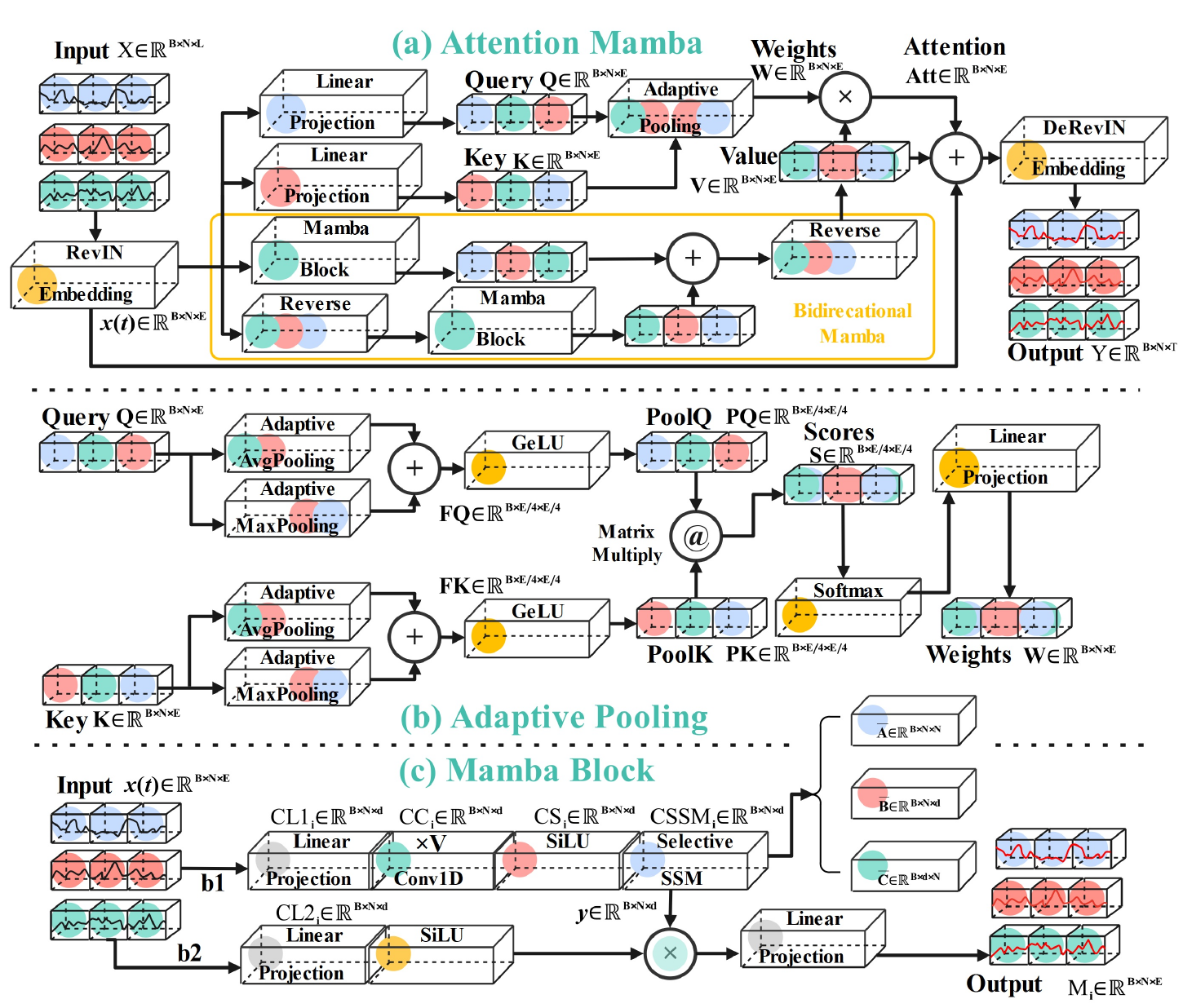}
\caption{The overall and detailed component architectures involved in Attention Mamba; (a) the overall architecture of Attention Mamba with paired RevIN layers used to reduce non-stationary occurrence; (b) the Adaptive Pooling block that accelerates attention computation and provides wider receptive fields; (c) the framework of Mamba that composes the core of the bidirectional Mamba block.}
\label{figure1}
\end{figure*}
The overall architecture of Attention Mamba is presented in Fig. \ref{figure1}(a), and the details of Adaptive Pooling block and Mamba block is depicted in Fig. \ref{figure1}(b) and Fig. \ref{figure1}(c) respectively. Attention Mamba employs two linear projections to obtain Query and Key at first, and then leverages the Adaptive Pooling block to compute Weights, while using the bidirectional Mamba block to compute the refined Value. Lastly, Attention Mamba empowers features with different weightings based on Weights and outputs the forecast sequence.

\subsection{Adaptive Pooling Acceleration}
To enhance the receptive fields of Attention Mamba and accelerate computational speed, we define a fused adaptive pooling equation as:
\begin{equation}
\label{eq:AP1}
\begin{aligned}
\mathbf{FQ} & = \operatorname{AvgPooling}(\mathbf{Q}) + \operatorname{MaxPooling}(\mathbf{Q}) \\
\mathbf{FK} & =\operatorname{AvgPooling}(\mathbf{K}) + \operatorname{MaxPooling}(\mathbf{K}),
\end{aligned}
\end{equation}
where $\mathbf{Q} \in \mathbb{R}^{B \times N \times E}$, $\mathbf{K} \in \mathbb{R}^{B \times N \times E}$ represents Query and Key respectively, $B$ represents the batch size, $N$ represents the number of variates, $E$ represents the dimension of embeddings, $\operatorname{AvgPooling}$ and $\operatorname{MaxPooling}$ are both adaptive and downscale the last two dimensions of Query and Key to a one quarter level, and $\mathbf{FQ} \in \mathbb{R}^{B \times E/4 \times E/4}$, $\mathbf{FK} \in \mathbb{R}^{B \times E/4 \times E/4}$ represents the Adaptive Pooling fused Query and Key, which both contain global features.

This receptive field bottleneck is resolved, and Adaptive Pooling in Eq. (\ref{eq:AP1}) simultaneously enhances nonlinear dependencies to Attention Mamba as adaptive pooling operation is generally recognized as nonlinear networks \cite{B014, B015}. We further enhance nonlinearity by adopting GeLU activation function \cite{B017} to obtain $\mathbf{PoolQ} \in \mathbb{R}^{B \times E/4 \times E/4}$ and $\mathbf{PoolK} \in \mathbb{R}^{B \times E/4 \times E/4}$, and compute $\mathbf{Scores}\in \mathbb{R}^{B \times E/4 \times E/4}$:
\begin{equation}
\label{eq:AP2}
\begin{aligned}
\mathbf{PoolQ} & = \operatorname{GeLU}(\mathbf{FQ}) \\
\mathbf{PoolK} & =\operatorname{GeLU}(\mathbf{FK}) \\
\mathbf{Scores} & = \mathbf{PoolQ} @\mathbf{PoolK},
\end{aligned}
\end{equation}
where $@$ denotes matrix multiplications.

Then, we activate and recover the last two dimensions back to obtain $\mathbf{Weights}\in \mathbb{R}^{B \times N \times E}$ \cite{B018}:
\begin{equation}
\label{eq:AP3}
\mathbf{Weights} = \operatorname{Linear Projection}(\operatorname{Softmax}(\mathbf{Scores})).
\end{equation}
\subsection{Attention Mechanism with Bidirectional Mamba}
In this project, we adopt a bidirectional Mamba block (\textbf{more details of Mamba can be seen in \cite{B5} and Fig. \ref{figure1}(c)}) to transform normalized inputs into $\mathbf{Value}\in \mathbb{R}^{B \times N \times E}$, and the bidirectional Mamba block is defined as \cite{B3}:
\begin{equation}
\label{eq:AM1}
\begin{aligned}
&\operatorname{Normal Process:}\quad\operatorname{Mamba}(x) \\
&\operatorname{Reverse Process:}\quad\operatorname{Mamba}(\operatorname{RVS}(x)) \\
&\mathbf{Value}=\operatorname{RVS}(\operatorname{Normal Process} + \operatorname{Reverse Process}),
\end{aligned}
\end{equation}
where $\operatorname{RVS}$ represents the reverse operation on sequences, $\operatorname{Mamba}$ represents all the process demonstrated in Fig. \ref{figure1}(c).

Lastly, we finalize the attention mechanism by:
\begin{equation}
\label{eq:AM2}
\mathbf{Att}=\mathbf{Weights} \times \mathbf{Value},
\end{equation}
where $\mathbf{Att} \in \mathbb{R}^{B \times N \times E}$.

\section{EXPERIMENTS}

\subsection{Experimental Setup}
To ensure the effectiveness of Attention Mamba across various real-world situations ranging from electricity consumptions, weather predictions, solar energy and traffic management, this project conducts experiments on 7 real-world datasets, whose characteristics are presented in Table.\ref{Datasets} \cite{B3}. These datasets involves different number of variates, sampling granularity, which are sufficient to guarantee the effectiveness of experiments.
\begin{table}[ht]
\centering
\caption{The Characteristics of Employed Datasets.}
\resizebox{\columnwidth}{!}{
\begin{tabular}{l|c|c|c}
\hline
\textbf{Datasets} & \textbf{Variates} & \textbf{Timesteps} & \textbf{Granularity} \\ \hline
Electricity       & 321               & 26,304             & 1hour                \\ \hline
Weather           & 21                & 52,696             & 10min                \\ \hline
Solar-Energy      & 137               & 52,560             & 10min                \\ \hline
PEMS03            & 358               & 26,209             & 5min                 \\ \hline
PEMS04            & 307               & 16,992             & 5min                 \\ \hline
PEMS07            & 883               & 28,224             & 5min                 \\ \hline
PEMS08            & 170               & 17,856             & 5min                 \\ \hline
\end{tabular}}
\label{Datasets}
\end{table}

To solidify the position of Attention Mamba among excellent counterparts and check whether Attention Mamba overcomes drawbacks of Transformer-based models,  Linear-based models, Mamba-based models, 10 representative SOTA models are cherry-picked as the benchmark. \textbf{Transformer-based models} include: iTransformer \cite{B8}, PatchTST \cite{B18}, Crossformer  \cite{B19}, FEDformer \cite{B20}, Autoformer \cite{B15}; \textbf{Linear-based models} include: RLinear  \cite{B21}, TiDE \cite{B22}, DLinear \cite{B23}; \textbf{Temporal Convolutional Network-based models} include: TimesNet \cite{B24}; \textbf{Mamba-based models} include: S-Mamba \cite{B3}. Moreover, we choose Mean Squared Error (MSE) and Mean Absolute Error (MAE) as the evaluation measurements.

Moreover, we present the default configuration of Attention Mamba as fo. The initial embedding dimension $E$ operates in a space $\mathbb{D} \in [32, 128, 256, 512]$. Each Mamba block has the expansion factor set to $1$ ($EF=1 \times \mathbb{D}$). The kernel size of Mamba is set to $KS\in[32,64, 128]$. The random seed is fixed at 2024, and optimization is performed using the Adam Optimizer. The initial learning rate is selected from the range $\in [1e-2, 1e-4]$. 

All the experiments are implemented on Windows with an NVIDIA GeForce RTX 4090 GPU. To enhance confidence in the findings, all results reported are obtained by averaging operations across five independent replications.

\section{RESULTS AND PERFORMANCE ANALYSIS}
\begin{table*}[tb]
\large
\centering
\caption{The comparisons of Attention Mamba, S-Mamba, iTransformer, RLinear, PatchTST, Crossformer, TiDE, TimeNet, DLinear, FEDformer, Autoformer across 7 employed datasets. The look-back window $L$ is set to 12 for PEMS and 96 for others. The forecast window $T$ varies: PEMS $\in[12, 24, 48, 96]$, others $\in [96, 192, 336, 720]$. \textit{Avg} represents the average results across the four defined forecast windows.}
\label{table3}
\begin{threeparttable}
\resizebox{\textwidth}{!}{
\begin{tabular}{l|c|cc|cc|cc|cc|cc|cc|cc|cc|cc|cc|cc} \hline
\multicolumn{2}{c|}{Models}          & \multicolumn{2}{c|}{Attention Mamba} & \multicolumn{2}{c|}{S-Mamba} & \multicolumn{2}{c|}{iTransformer} & \multicolumn{2}{c|}{RLinear} & \multicolumn{2}{c|}{PatchTST} & \multicolumn{2}{c|}{Crossformer} & \multicolumn{2}{c|}{TiDE} & \multicolumn{2}{c|}{TimesNet} & \multicolumn{2}{c|}{DLinear} & \multicolumn{2}{c|}{FEDformer} & \multicolumn{2}{c}{Autoformer} \\\hline
\multicolumn{2}{c|}{Metric}          & MSE              & MAE              & MSE          & MAE          & MSE             & MAE            & MSE          & MAE          & MSE           & MAE          & MSE            & MAE            & MSE         & MAE        & MSE           & MAE          & MSE          & MAE          & MSE           & MAE           & MSE            & MAE           \\\hline
\multirow{5}{*}{Electricity}  
    & 96  & \color{red}\textbf{0.138} & \color{red}\textbf{0.233} & \color{violet}\textbf{0.139} & \color{violet}\textbf{0.235} & 0.148 & 0.240 & 0.201 & 0.281 & 0.181 & 0.270 & 0.219 & 0.314 & 0.237 & 0.329 & 0.168 & 0.272 & 0.197 & 0.282 & 0.193 & 0.308 & 0.201 & 0.317 \\
    & 192 & \color{red}\textbf{0.157} & \color{red}\textbf{0.252} & \color{violet}\textbf{0.159} & 0.255 & 0.162 & \color{violet}\textbf{0.253} & 0.201 & 0.283 & 0.188 & 0.274 & 0.231 & 0.322 & 0.236 & 0.330 & 0.184 & 0.289 & 0.196 & 0.285 & 0.201 & 0.315 & 0.222 & 0.334 \\
    & 336 & \color{red}\textbf{0.174} & \color{red}\textbf{0.269} & \color{violet}\textbf{0.176} & \color{violet}\textbf{0.272} & 0.178 & \color{red}\textbf{0.269} & 0.215 & 0.298 & 0.204 & 0.293 & 0.246 & 0.337 & 0.249 & 0.344 & 0.198 & 0.300 & 0.209 & 0.301 & 0.214 & 0.329 & 0.231 & 0.338 \\
    & 720 & \color{red}\textbf{0.197} & \color{red}\textbf{0.293} & \color{violet}\textbf{0.204} & \color{violet}\textbf{0.298} & 0.225 & 0.317 & 0.257 & 0.331 & 0.246 & 0.324 & 0.280 & 0.363 & 0.284 & 0.373 & 0.220 & 0.320 & 0.245 & 0.333 & 0.246 & 0.355 & 0.254 & 0.361 \\
    & Avg & \color{red}\textbf{0.167} & \color{red}\textbf{0.262} & \color{violet}\textbf{0.170} & \color{violet}\textbf{0.265} & 0.178 & 0.270 & 0.219 & 0.298 & 0.205 & 0.290 & 0.244 & 0.334 & 0.251 & 0.344 & 0.192 & 0.295 & 0.212 & 0.300 & 0.214 & 0.327 & 0.227 & 0.338 \\\hline

\multirow{5}{*}{Weather}      
    & 96  & \color{violet}\textbf{0.162} & \color{red}\textbf{0.209} & 0.165 & \color{violet}\textbf{0.210} & 0.174 & 0.214 & 0.192 & 0.232 & 0.177 & 0.218 & \color{red}\textbf{0.158} & 0.230 & 0.202 & 0.261 & 0.172 & 0.220 & 0.196 & 0.255 & 0.217 & 0.296 & 0.266 & 0.336 \\
    & 192 & \color{violet}\textbf{0.210} & \color{violet}\textbf{0.253} & 0.214 & \color{red}\textbf{0.252} & 0.221 & 0.254 & 0.240 & 0.271 & 0.225 & 0.259 & \color{red}\textbf{0.206} & 0.277 & 0.242 & 0.298 & 0.219 & 0.261 & 0.237 & 0.296 & 0.276 & 0.336 & 0.307 & 0.367 \\
    & 336 & \color{red}\textbf{0.268} & \color{red}\textbf{0.295} & 0.274 & 0.297 & 0.278 & \color{violet}\textbf{0.296} & 0.292 & 0.307 & 0.278 & 0.297 & \color{violet}\textbf{0.272} & 0.335 & 0.287 & 0.335 & 0.280 & 0.306 & 0.283 & 0.335 & 0.339 & 0.380 & 0.359 & 0.395 \\
    & 720 & \color{violet}\textbf{0.349} & 0.348 & 0.350 & \color{red}\textbf{0.345} & 0.358 & \color{violet}\textbf{0.347} & 0.364 & 0.353 & 0.354 & 0.348 & 0.398 & 0.418 & 0.351 & 0.386 & 0.365 & 0.359 & \color{red}\textbf{0.345} & 0.381 & 0.403 & 0.428 & 0.419 & 0.428 \\
    & Avg & \color{red}\textbf{0.247} & \color{red}\textbf{0.276} & \color{violet}\textbf{0.251} & \color{red}\textbf{0.276} & 0.258 & \color{violet}\textbf{0.278} & 0.272 & 0.291 & 0.259 & 0.281 & 0.259 & 0.315 & 0.271 & 0.320 & 0.259 & 0.287 & 0.265 & 0.317 & 0.309 & 0.360 & 0.338 & 0.382 \\\hline

\multirow{5}{*}{Solar-Energy} 
    & 96  & \color{red}\textbf{0.197} & \color{red}\textbf{0.235} & 0.205 & 0.244 & \color{violet}\textbf{0.203} & \color{violet}\textbf{0.237} & 0.322 & 0.339 & 0.234 & 0.286 & 0.310 & 0.331 & 0.312 & 0.399 & 0.250 & 0.292 & 0.290 & 0.378 & 0.242 & 0.342 & 0.884 & 0.711 \\
    & 192 & \color{red}\textbf{0.228} & \color{violet}\textbf{0.262} & 0.237 & 0.270 & \color{violet}\textbf{0.233} & \color{red}\textbf{0.261} & 0.359 & 0.356 & 0.267 & 0.310 & 0.734 & 0.725 & 0.339 & 0.416 & 0.296 & 0.318 & 0.320 & 0.398 & 0.285 & 0.380 & 0.834 & 0.692 \\
    & 336 & \color{red}\textbf{0.248} & \color{violet}\textbf{0.279} & \color{violet}\textbf{0.258} & 0.288 & \color{red}\textbf{0.248} & \color{red}\textbf{0.273} & 0.397 & 0.369 & 0.290 & 0.315 & 0.750 & 0.735 & 0.368 & 0.430 & 0.319 & 0.330 & 0.353 & 0.415 & 0.282 & 0.376 & 0.941 & 0.723 \\
    & 720 & \color{violet}\textbf{0.252} & \color{violet}\textbf{0.284} & 0.260 & 0.288 & \color{red}\textbf{0.249} & \color{red}\textbf{0.275} & 0.397 & 0.356 & 0.289 & 0.317 & 0.769 & 0.765 & 0.370 & 0.425 & 0.338 & 0.337 & 0.356 & 0.413 & 0.357 & 0.427 & 0.882 & 0.717 \\
    & Avg & \color{red}\textbf{0.231} & \color{violet}\textbf{0.265} & 0.240 & 0.273 & \color{violet}\textbf{0.233} & \color{red}\textbf{0.262} & 0.369 & 0.356 & 0.270 & 0.307 & 0.641 & 0.639 & 0.347 & 0.417 & 0.301 & 0.319 & 0.330 & 0.401 & 0.291 & 0.381 & 0.885 & 0.711 \\\hline
\multirow{5}{*}{PEMS03}  
    & 12  & \color{red}\textbf{0.064}  & \color{red}\textbf{0.166}  & \color{violet}\textbf{0.065}  & \color{violet}\textbf{0.169}  & 0.071  & 0.174  & 0.126  & 0.236  & 0.099  & 0.216  & 0.090  & 0.203  & 0.178  & 0.305  & 0.085  & 0.192  & 0.122  & 0.243  & 0.126  & 0.251  & 0.272  & 0.385 \\
    & 24  & \color{red}\textbf{0.082}  & \color{red}\textbf{0.188}  & \color{violet}\textbf{0.087}  & \color{violet}\textbf{0.196}  & 0.093  & 0.201  & 0.246  & 0.334  & 0.142  & 0.259  & 0.121  & 0.240  & 0.257  & 0.371  & 0.118  & 0.223  & 0.201  & 0.317  & 0.149  & 0.275  & 0.334  & 0.440 \\
    & 48  & \color{red}\textbf{0.120}  & \color{red}\textbf{0.228}  & 0.133  & 0.243  & \color{violet}\textbf{0.125}  & \color{violet}\textbf{0.236}  & 0.551  & 0.529  & 0.211  & 0.319  & 0.202  & 0.317  & 0.379  & 0.463  & 0.155  & 0.260  & 0.333  & 0.425  & 0.227  & 0.348  & 1.032  & 0.782 \\
    & 96  & \color{violet}\textbf{0.189}  & \color{violet}\textbf{0.294}  & 0.201  & 0.305  & \color{red}\textbf{0.164}  & \color{red}\textbf{0.275}  & 1.057  & 0.787  & 0.269  & 0.370  & 0.262  & 0.367  & 0.490  & 0.539  & 0.228  & 0.317  & 0.457  & 0.515  & 0.348  & 0.434  & 1.031  & 0.796 \\
    & Avg & \color{violet}\textbf{0.114}  & \color{red}\textbf{0.219}  & 0.122  & 0.228  & \color{red}\textbf{0.113}  & \color{violet}\textbf{0.221}  & 0.495  & 0.472  & 0.180  & 0.291  & 0.169  & 0.281  & 0.326  & 0.419  & 0.147  & 0.248  & 0.278  & 0.375  & 0.213  & 0.327  & 0.667  & 0.601 \\\hline

\multirow{5}{*}{PEMS04}  
    & 12  & \color{red}\textbf{0.070} & \color{red}\textbf{0.173} & \color{violet}\textbf{0.076} & \color{violet}\textbf{0.180} & 0.078 & 0.183 & 0.138 & 0.252 & 0.105 & 0.224 & 0.098 & 0.218 & 0.219 & 0.340 & 0.087 & 0.195 & 0.148 & 0.272 & 0.138 & 0.262 & 0.424 & 0.491 \\
    & 24  & \color{red}\textbf{0.080} & \color{red}\textbf{0.187} & \color{violet}\textbf{0.084} & \color{violet}\textbf{0.193} & 0.095 & 0.205 & 0.258 & 0.348 & 0.153 & 0.275 & 0.131 & 0.256 & 0.292 & 0.398 & 0.103 & 0.215 & 0.224 & 0.340 & 0.177 & 0.293 & 0.459 & 0.509 \\
    & 48  & \color{red}\textbf{0.093} & \color{red}\textbf{0.203} & \color{violet}\textbf{0.115} & \color{violet}\textbf{0.224} & 0.120 & 0.233 & 0.572 & 0.544 & 0.229 & 0.339 & 0.205 & 0.326 & 0.409 & 0.478 & 0.136 & 0.250 & 0.355 & 0.437 & 0.270 & 0.368 & 0.646 & 0.610 \\
    & 96  & \color{red}\textbf{0.111} & \color{red}\textbf{0.222} & \color{violet}\textbf{0.137} & \color{violet}\textbf{0.248} & 0.150 & 0.262 & 1.137 & 0.820 & 0.291 & 0.389 & 0.402 & 0.457 & 0.492 & 0.532 & 0.190 & 0.303 & 0.452 & 0.504 & 0.341 & 0.427 & 0.912 & 0.748 \\
    & Avg & \color{red}\textbf{0.089} & \color{red}\textbf{0.196} & \color{violet}\textbf{0.103} & \color{violet}\textbf{0.211} & 0.111 & 0.221 & 0.526 & 0.491 & 0.195 & 0.307 & 0.209 & 0.314 & 0.353 & 0.437 & 0.129 & 0.241 & 0.295 & 0.388 & 0.231 & 0.337 & 0.610 & 0.590 \\\hline
\multirow{5}{*}{PEMS07}  
    & 12  & \color{red}\textbf{0.057} & \color{red}\textbf{0.150} & \color{violet}\textbf{0.063} & \color{violet}\textbf{0.159} & 0.067 & 0.165 & 0.118 & 0.235 & 0.095 & 0.207 & 0.094 & 0.200 & 0.173 & 0.304 & 0.082 & 0.181 & 0.115 & 0.242 & 0.109 & 0.225 & 0.199 & 0.336 \\
    & 24  & \color{red}\textbf{0.072} & \color{red}\textbf{0.168} & \color{violet}\textbf{0.081} & \color{violet}\textbf{0.183} & 0.088 & 0.190 & 0.242 & 0.341 & 0.150 & 0.262 & 0.139 & 0.247 & 0.271 & 0.383 & 0.101 & 0.204 & 0.210 & 0.329 & 0.125 & 0.244 & 0.323 & 0.420 \\
    & 48  & \color{red}\textbf{0.085} & \color{red}\textbf{0.175} & \color{violet}\textbf{0.093} & \color{violet}\textbf{0.192} & 0.110 & 0.215 & 0.562 & 0.541 & 0.253 & 0.340 & 0.311 & 0.369 & 0.446 & 0.495 & 0.134 & 0.238 & 0.398 & 0.458 & 0.165 & 0.288 & 0.390 & 0.470 \\
    & 96  & \color{red}\textbf{0.102} & \color{red}\textbf{0.195} & \color{violet}\textbf{0.117} & \color{violet}\textbf{0.217} & 0.139 & 0.245 & 1.096 & 0.795 & 0.346 & 0.404 & 0.396 & 0.442 & 0.628 & 0.577 & 0.181 & 0.279 & 0.594 & 0.553 & 0.262 & 0.376 & 0.554 & 0.578 \\
    & Avg & \color{red}\textbf{0.079} & \color{red}\textbf{0.172} & \color{violet}\textbf{0.089} & \color{violet}\textbf{0.188} & 0.101 & 0.204 & 0.504 & 0.478 & 0.211 & 0.303 & 0.235 & 0.315 & 0.380 & 0.440 & 0.124 & 0.225 & 0.329 & 0.395 & 0.165 & 0.283 & 0.367 & 0.451 \\\hline
\multirow{5}{*}{PEMS08}  
    & 12  & \color{red}\textbf{0.075} & \color{red}\textbf{0.174} & \color{violet}\textbf{0.076} & \color{violet}\textbf{0.178} & 0.079 & 0.182 & 0.133 & 0.247 & 0.168 & 0.232 & 0.165 & 0.214 & 0.227 & 0.343 & 0.112 & 0.212 & 0.154 & 0.276 & 0.173 & 0.273 & 0.436 & 0.485 \\
    & 24  & \color{red}\textbf{0.098} & \color{red}\textbf{0.197} & \color{violet}\textbf{0.104} & \color{violet}\textbf{0.209} & 0.115 & 0.219 & 0.249 & 0.343 & 0.224 & 0.281 & 0.215 & 0.260 & 0.318 & 0.409 & 0.141 & 0.238 & 0.248 & 0.353 & 0.210 & 0.301 & 0.467 & 0.502 \\
    & 48  & \color{red}\textbf{0.139} & 0.237 & \color{violet}\textbf{0.167} & \color{red}\textbf{0.228} & 0.186 & \color{violet}\textbf{0.235} & 0.569 & 0.544 & 0.321 & 0.354 & 0.315 & 0.355 & 0.497 & 0.510 & 0.198 & 0.283 & 0.440 & 0.470 & 0.320 & 0.394 & 0.966 & 0.733 \\
    & 96  & 0.264 & 0.298 & \color{violet}\textbf{0.245} & \color{violet}\textbf{0.280} & \color{red}\textbf{0.221} & \color{red}\textbf{0.267} & 1.166 & 0.814 & 0.408 & 0.417 & 0.377 & 0.397 & 0.721 & 0.592 & 0.320 & 0.351 & 0.674 & 0.565 & 0.442 & 0.465 & 1.385 & 0.915 \\
    & Avg & \color{red}\textbf{0.144} & 0.227 & \color{violet}\textbf{0.148} & \color{red}\textbf{0.224} & 0.150 & \color{violet}\textbf{0.226} & 0.529 & 0.487 & 0.280 & 0.321 & 0.268 & 0.307 & 0.441 & 0.464 & 0.193 & 0.271 & 0.379 & 0.416 & 0.286 & 0.358 & 0.814 & 0.659 \\\hline
\end{tabular}}
\label{Results1}
\begin{tablenotes}
\footnotesize
\item \text{The \textbf{\color{red}Best}/\textbf{\color{violet}Second Best} results are indicated by \text{\textbf{Bold}/\color{red}{Red}}, \textbf{Bold}/\color{violet}{Violet}.}
\end{tablenotes}
\end{threeparttable}
\end{table*}
\subsection{Nonlinear Dependency Extraction Effectiveness}
\label{Nonlinear Dependency Extraction Effectiveness}
To validate the nonlinear dependency extraction effectiveness, as shown in Table. \ref{Results1}, the leading positions used to be possessed by S-Mamba and iTransformer. \textbf{In fact, S-Mamba is a bidirectional Mamba model, and iTransformer contains attention mechanisms. The comparisons between Attention Mamba and these two models can demonstrate the effectiveness of adaptive pooling techniques plus the attention mechanism, and the bidirectional Mamba respectively. To some extent, the mentioned comparisons are ablation studies as well.} 

To be specific, in Electricity, Attention Mamba beats both S-Mamba and iTransformer across all forecasting windows. However, there are minor gaps between Attention Mamba and S-Mamba, indicating that the common bidirectional Mamba mechanism is the driving factor to achieve similar performances. Another reason is that we set the initial embedding $E$ to $512$, larger than the number of variates of Electricity at 321 as shown in Table. \ref{Datasets}, providing sufficient receptive fields for Attention Mamba and shedding the light of the Adaptive Pooling block.

Moreover, in Weather, it is obvious that Attention Mamba and S-Mamba share similar results in MAE measurements. The main explanation is that both Attention Mamba and S-Mamba take MSE-oriented training processes, leading to clear differences in MSE. Furthermore, although Attention Mamba does not achieve SOTA under all forecasting windows, the comprehensive performance of Attention Mamba achieves the best, as leading results of different forecasting windows scatter between Attention Mamba, Crossformer, and DLinear. This indicates that Attention Mamba has competitiveness in small situations like Weather, compared with Transformer-based models and Linear-based models.

Additionally, in Solar Energy, it is difficult to identify whether Attention Mamba or iTransformer holds the first place, as Attention Mamba outperforms iTransformer if measured by MSE, while Attention Mamba underperforms iTransformer in terms of MAE. We will further conduct Friedman nonparametric testing in Section. \ref{ranking performance analysis} to compare the comprehensive performance of all models. However, the comparison between Attention Mamba and S-Mamba confirms that the Adaptive Pooling block can enhance nonlinear dependency extractions, and Table. \ref{Solar Comparison} summarizes such enhancements. As shown in Table. \ref{Solar Comparison}, the Adaptive Pooling blocks can generate around $3\%$ improvements in S-Mamba's performance, confirming the effectiveness of Adaptive Pooling techniques.

In summary, the overall performance of Attention Mamba in Electricity, Weather, and Solar Energy datasets draws three important conclusions: 
\begin{itemize}
    \item The Adaptive Pooling block is fixed to reduce the last two dimensions of sequences to one quarter of the initial embedding $E$ and becomes less effective in certain situations, such as Weather datasets. A dynamic input-dependent scheme can be developed to break such limitations.
    \item Both the bidirectional Mamba block and the Adaptive Pooling block can enhance nonlinear dependency extractions, leading to comprehensive performance improvements.
    \item Attention Mamba grows to be competitive to iTransformer (Transformer-based models), DLinear (Linear-based models), and S-Mamba (Mamba-based models).
\end{itemize}

\begin{table}[tb]
\centering
\caption{The improvement ratios of Attention Mamba compared with S-Mamba in MSE and MAE in Solar Energy datasets.}
\resizebox{\columnwidth}{!}{
\begin{tabular}{l|c|c|c|c|c}
\hline
\textbf{Ratio} & \textbf{96} & \textbf{192} & \textbf{336} & \textbf{720} & \textbf{Avg}\\ \hline
\textbf{MSE}    & 3.90\%                 & 3.80\%             & 3.88\%                & 3.08\%           & 3.75\%      \\ \hline
\textbf{MAE}    & 3.69\%                 & 2.96\%             & 3.13\%                & 1.39\%           & 2.93\%      \\ \hline
\end{tabular}}
\label{Solar Comparison}
\end{table}

\subsection{Receptive Field Enhancements}

Apart from nonlinear dependency extraction improvements, the performance of Attention Mamba in PEMS03 to PEMS08 also confirms the effectiveness of the Adaptive Pooling block in receptive field enhancements. As we set the initial embedding $E$ to $256$ and $512$ accordingly, smaller than the numbers of variates range from $307$ to $883$. As shown in Table. \ref{Results1}, Attention Mamba significantly outperforms S-Mamba, securing the first position, and the improvement ratios are summarized in Table. \ref{PEMS Comparision}. 

\begin{table}[tb]
\centering
\caption{The improvement ratios of Attention Mamba compared with S-Mamba in average MSE and MAE in PEMS datasets.}
\resizebox{\columnwidth}{!}{
\begin{tabular}{l|c|c|c|c}
\hline
\textbf{Ratio} & \textbf{PEMS03} & \textbf{PEMS04} & \textbf{PEMS07} & \textbf{PEMS08}\\ \hline
\textbf{MSE}    & 7.14\%                 & 13.59\%             & 11.24\%                & 2.7\%                \\ \hline
\textbf{MAE}    & 3.94\%                 & 7.11\%             & 8.51\%                & -1.34\%                \\ \hline
\end{tabular}}
\label{PEMS Comparision}
\end{table}

As shown in Table. \ref{PEMS Comparision}, notable improvements exceeding $3\%$ (this enhancement is attributed to the extraction of nonlinear dependencies.) are observed in PEMS03, PEMS04, and PEMS07. The surplus enhancements (e.g., $(13.59-3)\%$) are considered to be generated from the enhanced receptive fields of Attention Mamba, as the Adaptive Pooling block provides global information and insights to all the embeddings, variate channels.

Moreover, as shown in Table. \ref{PEMS Comparision}, there are an approximate $3\%$ ($2.70\%$) in MSE and a sharp deterioration in MAE at $-1.34\%$ in terms of PEMS08. The $2.70\%$ confirms our conclusion in Section. \ref{Nonlinear Dependency Extraction Effectiveness} that this enhancement should be attributed to nonlinear dependency extraction enhancements, because the number of variates in PEMS08 is $170$ as shown in Table. \ref{Datasets}, which is smaller than the lowest embedding dimension at $256$. The deterioration also complies with our conclusion that the MSE-oriented training process can results in negative impacts on MAE measurements.

\begin{figure}[tb]
	\centering 
	\includegraphics[width=0.5\textwidth, angle=0]{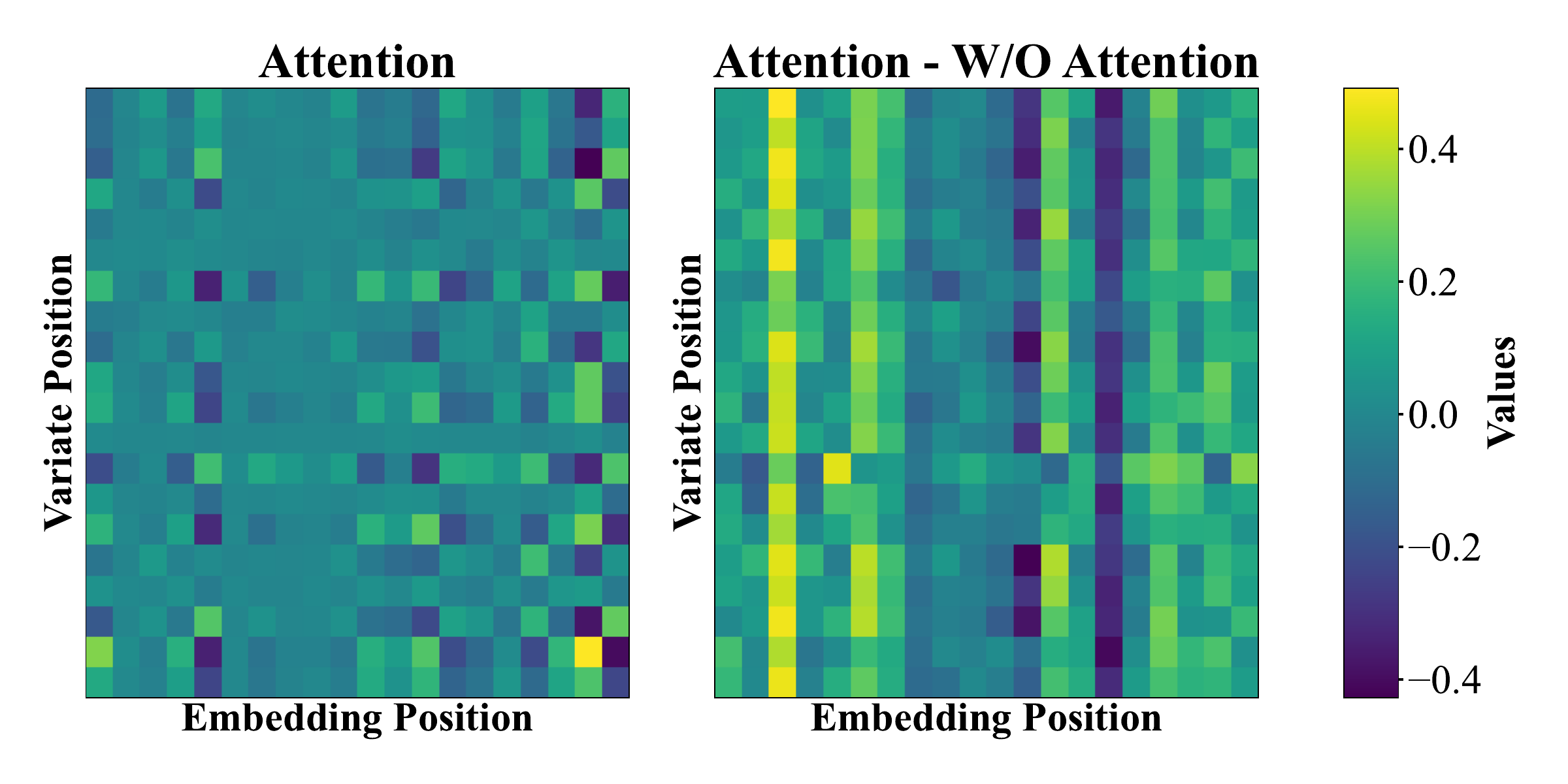}	
	\caption{A snapshot of the attention map of PEMS04 with the forecasting window at 48 to visually demonstrate the enhanced receptive fields. Attention represents the input weighted by the Adaptive Pooling block, and W/O Attention represents the original input.} 
	\label{fig:attention map}%
\end{figure}

Furthermore, a visualized snapshot is displayed in Fig. \ref{fig:attention map} and reveals that the Adaptive Pooling block can significantly enhance receptive fields and pay attention to key features.

In summary, the Adaptive Pooling block can adequate provide global information and significantly enhance the receptive fields of Attention Mamba, leading to considerable improvements in accuracy.

\subsection{Comprehensive Ranking Performance Analysis}
\label{ranking performance analysis}
\begin{table*}[tb]
\centering
\caption{The Friedman Ranking of Attention Mamba and all benchmarks. The insignificant results ($p$-value$>0.05$) are marked in \text{\color{red}red text}.}
\label{Friedman Ranking Benchmark}
\begin{adjustbox}{max width=\textwidth}
\begin{tabular}{lcccccccc}
\toprule
\multirow{2}{*}{Model} & \multicolumn{4}{c}{MSE} & \multicolumn{4}{c}{MAE} \\
\cmidrule(lr){2-5} \cmidrule(lr){6-9}
 & Rank & $z$-value & unadjusted $p$ & $p_{\text{Bonf}}$ & Rank & $z$-value & unadjusted $p$ & $p_{\text{Bonf}}$ \\
\midrule
Autoformer & 10.39 & 11.532 & 0 & 0 & 10.61 & 11.568037 & 0 & 0 \\
Crossformer & 6.54 & 6.684956 & 0 & 0 & 6.94 & 6.937218 & 0 & 0 \\
FEDformer & 6.91 & 7.153443 & 0 & 0 & 7.53 & 7.675987 & 0 & 0 \\
iTransformer & 2.89 & 2.072156 & 0.038251 & 0.038251 & 2.39 & 1.189237 & \text{\color{red}0.234346} & \text{\color{red}0.234346} \\
PatchTST & 5.63 & 5.531756 & 0 & 0 & 5.09 & 4.594781 & 0.000004 & 0.000004 \\
TimeNet & 4.43 & 4.018181 & 0.000059 & 0.000059 & 4.51 & 3.874031 & 0.000107 & 0.000107 \\
TiDE & 9.29 & 10.144556 & 0 & 0 & 9.46 & 10.108518 & 0 & 0 \\
RLinear & 9.09 & 9.892293 & 0 & 0 & 7.91 & 8.162493 & 0 & 0 \\
DLinear & 7.26 & 7.585893 & 0 & 0 & 7.91 & 8.162493 & 0 & 0 \\
S-Mamba & 2.34 & 1.387444 & \text{\color{red}0.165307} & \text{\color{red}0.165307} & 2.20 & 0.954994 & \text{\color{red}0.339581} & \text{\color{red}0.339581} \\
Attention Mamba & 1.24 & - & - & - & 1.44 & - & - & - \\
\bottomrule
\end{tabular}
\end{adjustbox}
\end{table*}
To solidify the positions of Attention Mamba among leading counterparts and confirms our previous conclusions, in this section, we conduct Friedman nonparametric testing to rank all the models and deliver a direct ranking in Table. \ref{Friedman Ranking Benchmark}.

As shown in Table. \ref{Friedman Ranking Benchmark}, Attention Mamba ranks the first in both MSE and MAE. However, not all results are significant, and under MSE-oriented training processes, Attention Mamba beats all the counterparts, except for S-Mamba, as the results of S-Mamba is insignificant, indicating that S-Mamba and Attention Mamba performs nearly the same across some datasets. Given the results in Table. \ref{Results1}, we consider this insignificance is caused by similar performance in Electricity, Weather, and PEMS08. After comprehensively considerations, we consider Attention Mamba competitive to leading models such as S-Mamba, iTransformer.

\subsection{Memory and Complexity}
Given the enhancements in performance can be generated from an exchange of complexity (compared to a bidirectional Mamba model, S-Mamba), herein, we compare the cost in memory and complexity under PEMS07 (the biggest dataset for clear observations) with the lookback window length at $12$ for the top five models. The comparison is presented in Table. \ref{Memory and Compolexity}.
\begin{table}[!htbp]
\centering
\caption{The Comparison of Attention Mamba in Training Time (ms/iteration) and Memory Occupation (Gb) with Other Four Top Models. (-) Represents the Change Ratios Compared to Attention Mamba. \text{\color{red}RED} Represents A Better Performance.}
\resizebox{\columnwidth}{!}{
\begin{tabular}{l|c|c|c}
\hline
\textbf{Values} & \textbf{Training Time (Change)} & \textbf{Memory (Change)} & \textbf{MSE (Change)} \\ \hline
\textbf{RLinear}&59.20 (\text{\color{red}-14.3\%}) &0.028 (\text{\color{red}-98.2\%}) &0.118 (+107.0\%)\\ \hline
\textbf{PatchTST}&575.71 (+733.3\%)&1.570 (+0.64\%) &0.095 (+66.7\%)\\ \hline
\textbf{Crossformer}&309.28 (347.6\%) &1.830 (+17.3\%) &0.094 (+64.9\%)\\ \hline
\textbf{iTransformer}&197.86 (+186.4\%) &1.670 (+7.1\%) &0.067 (+17.5\%)\\ \hline
\textbf{S-Mamba}&98.51 (+42.6\%)&1.030 (\text{\color{red}-34.0\%}) &0.063 (+10.5\%)\\ \hline
\textbf{Attention Mamba}&69.09 (-) &1.560 (-) &0.057 (-)\\ \hline
\end{tabular}}
\label{Memory and Compolexity}
\end{table}

From Table. \ref{Memory and Compolexity}, it clearly demonstrates that compared with RLinear (Linear-Based), Attention Mamba reaches a doubled accuracy at the cost of nearly doubled memory occupations and a slightly slower training speed. However, compared with S-Mamba (Bidirectional Mamba), Attention Mamba has advantages in training time, which is reduced by 42.6\%, while the memory occupation rises due to the attention mechanism by 34\%. It is interesting that Attention Mamba achieves 10.5\% enhancements in accuracy, which comprehensively comes from $42.6\%-34.0\%=8.6\%<10.5\%$. Therefore, we consider the benefits slightly outweigh the costs.

In summary, we consider Attention Mamba does not increase too much training time and memory, and has an excellent performance in these two perspectives.

\section{CONCLUSIONS}

In essence, this work addresses limitations in existing representative models by proposing Attention Mamba. The insufficiency of attention and limited receptive fields are primarily overcome by designing the Adaptive Pooling block that strengthens nonlinear dependencies while providing global information, leading to a widened receptive fields. While the Adaptive Pooling block's effectiveness may be constrained when the embedding dimension does not align with the number of dataset variables, extensive experiments demonstrate that Attention Mamba consistently achieves SOTA performance across employed benchmarking datasets, outperforming leading counterparts. Besides, the benefits obtained in accuracy almost are equal to or outweigh the costs generated from training time and memory occupations. In conclusion, the Adaptive Pooling block effectively enhances the receptive fields and ability to extract nonlinear dependencies of bidirectional Mamba mechanisms. Looking towards future, a dynamic input-dependent adaptive pooling scheme can be developed to alleviate current bottlenecks.

\section*{ACKNOWLEDGMENT}
This work was supported by JST CREST Grant Number JPMJCR22D1 and JSPS KAKENHI Grant Number JP22H00551, Japan.

\bibliographystyle{IEEEbib}
\bibliography{refs}

\end{document}